\documentclass[journal]{IEEEtran}
\usepackage[T1]{fontenc}
\usepackage{graphicx}
\usepackage{times}
\usepackage{helvet}
\usepackage{courier}
\usepackage{amsmath}
\usepackage{algorithm}
\usepackage{algorithmic}
\usepackage{csquotes} 
\usepackage{color}
\usepackage{paralist}
\usepackage{amssymb}
\usepackage{indentfirst}
\usepackage{subfigure}
\usepackage{float}
\usepackage{multirow}
\usepackage{cite}
\usepackage{mathrsfs}

\usepackage{mathrsfs} 
\usepackage{amsfonts}
\usepackage{todonotes}
\usepackage{pgfplots} 
\usepackage{epstopdf}
\usepackage{epsfig}
\pgfplotsset{compat=newest}
\usepackage{color}
\definecolor{forestgreen}{RGB}{0,139,69}

\usepackage{xcolor}
\definecolor{citecolor}{HTML}{0071bc}
\usepackage[colorlinks, linkcolor=red,  anchorcolor=blue, citecolor=citecolor]{hyperref} 

\usepackage{xcolor}
\definecolor{SeaGreen4}{RGB}{0,205,102} 
\definecolor{SlateBlue}{RGB}{106,90,205} 
\definecolor{DarkRed}{RGB}{178,34,34} 
	
\usepackage[switch]{lineno}

\usepackage{textcomp,booktabs}
\usepackage{amssymb}
\usepackage{pifont}

\usepackage{makecell}

\usepackage{colortbl}
\definecolor{mygray}{gray}{.9}
\definecolor{mypink}{rgb}{.99,.91,.95}
\definecolor{mycyan}{cmyk}{.3,0,0,0}

\begin{document}

\title{XiHeFusion: Harnessing Large Language Models for Science Communication in Nuclear Fusion}   

\author{Xiao Wang, Qingquan Yang*, Fuling Wang, Qiang Chen, Wentao Wu, Yu Jin, Jingtao Jiang, Liye Jin, \\ 
    Bo Jiang, Dengdi Sun, Wanli Lv, Meiwen Chen, Zehua Chen, Guosheng Xu, Jin Tang* 

\thanks{ $\bullet$ Xiao Wang, Fuling Wang, Qiang Chen, Wentao Wu, Yu Jin, Jingtao Jiang, Liye Jin, 
    Bo Jiang, Dengdi Sun, Wanli Lv, and Meiwen Chen are with Anhui University, Hefei 230601, China. (email: xiaowang@ahu.edu.cn)} 
\thanks{$\bullet$ Qingquan Yang, Guosheng Xu are with the Institute of Plasma Physics, Chinese Academy of Sciences, Hefei, China.}  
\thanks{$\bullet$ Zehua Chen is with the Department of Computer Science and Technology, Tsinghua University, Beijing
100190, China.}  

\thanks{* Corresponding Author: Qingquan Yang, Jin Tang (email: yangqq@ipp.ac.cn, tangjin@ahu.edu.cn)}   
}

\markboth{XiHeFusion: Harnessing Large Language Models for Science Communication in Nuclear Fusion} 
{Shell \MakeLowercase{\textit{et al.}}: Bare Demo of IEEEtran.cls for IEEE Journals}

\maketitle

\begin{abstract}
Nuclear fusion is one of the most promising ways for humans to obtain infinite energy. Currently, with the rapid development of artificial intelligence, the mission of nuclear fusion has also entered a critical period of its development. How to let more people to understand nuclear fusion and join in its research is one of the effective means to accelerate the implementation of fusion. This paper proposes the first large model in the field of nuclear fusion, XiHeFusion, which is obtained through supervised fine-tuning based on the open-source large model Qwen2.5-14B. We have collected multi-source knowledge about nuclear fusion tasks to support the training of this model, including the common crawl, eBooks, arXiv, dissertation, etc. After the model has mastered the knowledge of the nuclear fusion field, we further used the chain of thought to enhance its logical reasoning ability, making XiHeFusion able to provide more accurate and logical answers. In addition, we propose a test questionnaire containing 180+ questions to assess the conversational ability of this science popularization large model. Extensive experimental results show that our nuclear fusion dialogue model, XiHeFusion, can perform well in answering science popularization knowledge. 
The pre-trained XiHeFusion model is released on \url{https://github.com/Event-AHU/XiHeFusion}. 
\end{abstract}


\IEEEpeerreviewmaketitle

\section{Introduction}

\IEEEPARstart{A}{lthough} there are already various forms of energy such as solar, wind, coal, oil, and natural gas, energy issues have always been one of the key problems troubling humanity, such as long renewable cycles and severe environmental pollution. With the rapid development of physics, humans have mastered nuclear energy and successfully applied nuclear fission technology to power generation. However, nuclear fission easily produces waste with nuclear radiation, and the raw materials are expensive, therefore, nuclear fission is not an ideal future energy source. Nuclear fusion offers several key advantages over nuclear fission, e.g., abundant fuel, high energy yield, reduced waste, environmental safety, inherent safety, and non-proliferation. Despite these benefits, technical hurdles remain, including achieving and maintaining the extreme conditions for fusion and efficiently converting fusion energy into electricity.

To address these challenges, many countries around the world have established or are constructing nuclear fusion devices to explore this future energy source. Specifically, China has built the EAST large scientific facility, the United States has constructed the DIII-D, the European Union has established JET, and there is the multi-nationally constructed ITER facility, among others. Currently, nuclear fusion research is still primarily focused on scientific experimentation and physical model design. Although significant progress has been made in the past, there is still a long way to go before achieving a truly positive energy output.

In order to help more people understand nuclear fusion, especially the basic concepts, and working principles, and to enable newcomers to get up to speed in this field more quickly, this paper proposes a novel conversational large language model for nuclear fusion, termed \textbf{XiHeFusion}. To pre-train this large language model, we collected multi-sourced knowledge on nuclear fusion as shown in Table~\ref{tab:dataset}, including CommonCrawl, CNKI (China National Knowledge Infrastructure), eBooks, arXiv, and dissertation. We then used the large model DeepSeek V3~\cite{deepseekai2024deepseekv3} to process this information into more than 1 million question-answer pairs (about 370 million tokens), which served as the corpus for training the large model. We conducted supervised fine-tuning on a foundation model Qwen2.5-14B~\cite{yang2024qwen2.5}. To enhance the model's reasoning capabilities and provide more detailed and logical responses, we further explored the Chain-of-Thought (CoT)~\cite{wei2022CoT} technique to improve the model's question-answering abilities. Additionally, we invited domain experts to prepare test questionnaires which contain 184 questions to assess the question-answering capabilities of the XiHeFusion, as shown in Fig.~\ref{fig:NFAssessment}.

The features of our proposed XiHeFusion can be summarized as follows: 

\noindent $\bullet$ \textbf{[First Nuclear Fusion LLM]} It is the first large language model developed for the plasma nuclear fusion domain, effectively supporting science popularization in nuclear fusion to enhance the public's understanding of this field. 

\noindent $\bullet$ \textbf{[Open Source \& Bilingual Dialogue]} The XiHeFusion is fine-tuned based on open-source large model Qwen2.5-14B~\cite{yang2024qwen2.5}, which supports bilingual dialogue in both Chinese and English, and demonstrates strong generalization. 

\noindent $\bullet$ \textbf{[Fusion Knowledge-enhanced Training]} To enable the large language model to provide more professional responses to questions in the fusion field, we have collected a large-scale dataset from multiple sources to support self-supervised training. 

\noindent $\bullet$ \textbf{[Logical Dialogue]} The use of Chain-of-Thought (CoT) reasoning techniques ensures that the XiHeFusion large model can provide more detailed and logically thought-out answers. 

\noindent $\bullet$ \textbf{[New Test Questionnaire]} We have developed a science popularization quiz on nuclear fusion, which examines fusion knowledge from multiple perspectives. It can effectively test the large model's mastery of domain knowledge.

\textit{\textbf{The rest of this paper is organized as follows}}: 
We introduce the related works on the Large Language Model, Nuclear Fusion, and Chain-of-Thought in Section~\ref{relatedWorks}. After that, we introduce the XiHeFusion large language model in Section~\ref{sec::XiHeFusion}, with a focus on data collection and pre-processing, network architecture, and optimization. The introduced questions for the evaluation are described in Section~\ref{sec::NFAssessment}. We introduce the experiments in Section~\ref{sec::Experiments} and focus on comparing XiHeFusion with other large language models, visualization and analysis of question-answer cases, and limitation analysis. We conclude this paper in Section~\ref{sec::conclusion}.

\section{Related Works}  \label{relatedWorks}

In this section, we will review the related works on the Large Language Model, Nuclear Fusion, and Chain-of-Thought. More related works can be found in the following surveys~\cite{wang2023MMPTMs} and paper list\footnote{\url{https://github.com/Event-AHU/AI_for_Controllable_Nuclear_Fusion/blob/main/Survey_Paper_list.md}}.

\subsection{Large Language Model}  
LLMs have demonstrated remarkable language understanding and the ability to handle complex tasks through text generation~\cite{jin2024MSP60K, wang2024r2gencsr, wang2025AMMRG}. 
More in detail, GPT-3.0~\cite{kojima2022letsthinkstepbystep}, developed by OpenAI, was the first large language model to achieve industrial success, with 175 billion parameters enabling it to excel in natural language tasks. Its success spurred rapid advancements in large language models, leading to improved versions like GPT-4~\cite{achiam2023gpt4}, which offers stronger reasoning and broader knowledge. OpenAI o1\footnote{\url{https://openai.com/index/learning-to-reason-with-llms/}} gained attention for its exceptional complex reasoning, leveraging reinforcement learning and chain-of-thought training to surpass human PhD-level performance on the GPQA benchmark~\cite{rein2023gpqa} for physics, biology, and chemistry. LLaMA~\cite{touvron2023llama1} adopts a \textit{small models, large data} approach, producing high-performance models. Llama-1~\cite{touvron2023llama1}, offers four parameter sizes: 7B, 13B, 30B, and 65B, was trained on 1T+ tokens, while Llama-2~\cite{touvron2023llama2} expanded to 2T tokens, doubled context length to 4,096, and introduced GQA. Llama-3~\cite{grattafiori2024llama3} supports 8K contexts, uses a 128K vocabulary, and trains on over 15T tokens, delivering state-of-the-art performance with improved inference, code generation, and instruction-following capabilities. Gemini~\cite{team2023gemini}, Google's most advanced AI model, comes in three versions (Ultra, Pro, Nano) and supports diverse scenarios, focusing on complex reasoning, multimodal understanding, and coding. Claude\footnote{\url{https://claude.ai}}, developed by Anthropic, is a GPT-like AI model prioritizing safety, reliability, and alignment, with multiple improved versions released.

On the other hand, Qwen~\cite{bai2023qwen} has consistently focused on the technical development of foundational models, advancing from its initial version to the latest 2.5 release. Compared to the previous version, the Qwen2.5~\cite{yang2024qwen2.5} demonstrates significant improvements in comprehension, logical reasoning, instruction following, and coding capabilities, with its Chinese language proficiency continuing to lead the industry. DeepSeek-V3~\cite{deepseekai2024deepseekv3} has 671 billion parameters, with 37 billion activated, offering performance on par with top models in knowledge-based Q\&A, long-text processing, code generation, and mathematical reasoning, while being more cost-efficient. The Spark LLM\footnote{\url{https://xinghuo.xfyun.cn/}} by iFlytek excels in natural language processing for customer service, education, and healthcare. Tiangong\footnote{\url{https://www.tiangong.cn/}} is China's first dual-trillion-parameter model, outperforming ChatGPT in tasks like content creation, logical reasoning, and mathematical computation, providing efficient support for intelligent search, recommendation systems, and virtual assistants. Other LLMs, such as Baichuan~\cite{yang2023baichuan2}, Ernie Bot~\cite{sun2019erniebot}, Doubao\footnote{\url{https://www.doubao.com/chat/}}, SenseChat\footnote{\url{https://chat.sensetime.com/}}, and Bing Chat\footnote{\url{https://copilot.microsoft.com/}}, each have their unique features, covering a wide range of capabilities from multi-modal processing and code generation to conversational interactions. They are driving the deep application of artificial intelligence in various fields and accelerating the iteration and innovation of technology.




\subsection{Nuclear Fusion}

With the advancement of nuclear fusion, deep learning has found increasing applications in nuclear fusion research, aiding in solving complex physical problems and optimizing experimental processes, such as Q-distribution prediction~\cite{wang2024MMQdist, ma2024MHNQdist}, plasma state prediction, Tokamak control optimization, and plasma diagnostics. 
Yamaguchi et al.~\cite{yamaguchi2021geneticoptimization} uses a genetic algorithm to optimize the control points of three-dimensional B-spline curves, to solve the problem of designing and optimizing external coils for stellarators. 
Hu et al.~\cite{hu2021randomforest} solve the problem of real-time disruption prediction and mitigation in high-density discharges of the EAST tokamak by developing a random forest-based real-time disruption predictor (DPRF), improving the accuracy of disruption alarms and reducing disruption damage. 
Schmidt et al.~\cite{schmidt2024neural} employ a deep convolutional neural network to reconstruct fast-ion velocity distributions from fast-ion loss detectors and imaging neutral particle analyzers (INPAs). PlaNet~\cite{bonotto2024planet} solves the problem of fast and accurate plasma equilibrium and separatrix reconstruction using a physics-informed deep learning approach. 
Inoue et al.~\cite{inoue2024svm} use a Support Vector Machine (SVM) combined with redundant logic and an adaptive voltage allocation scheme to mitigate the risks of asymmetric heat loads on the first wall and electromagnetic loads on conductive materials caused by Vertical Displacement Events (VDEs). 
SExFC~\cite{li2024RNNGRU} integrates the recurrent neural network (RNN) algorithm and utilizes the Gated Recurrent Unit (GRU) for iterative prediction of flux evolution based on radial profiles. 
Zhang et al.~\cite{zhang2024yolo} use YOLO (You Only Look Once)~\cite{cvpr2016yolo, cvpr2017yolo9000, farhadi2018yolov3} to identify Ion Cyclotron Emission (ICE) in HL-2A discharges, aiming to enhance real-time fast ion diagnostics for magneto hydro dynamics (MHD) instabilities in fusion plasmas. 
Sun et al.~\cite{sun2024impactmlp} develop a multi-layer perceptron (MLP) neural network model as a surrogate for kinetic equilibrium fitting (EFITs) and investigate the impact of different diagnostic data and machine actuator controls on the accuracy of equilibrium reconstruction. 
Wan et al.~\cite{wan2023transformer} applies a transformer-based model to solve the real-time reconstruction of the last closed flux surface (LCFS) in the experimental advanced superconducting tokamak (EAST).

Some researchers adopt CNNs~\cite{boyer2024cnn_neural, seo2024cnn_avoiding, lin2024cnn_prediction, zanisi2024cnn_efficient, joung2024cnn_mlp_tokamak, bonotto2024cnn_mlp_reconstruction}, MLPs~\cite{sun2024mlp_impact, sanchez2024mlp_real, mehta2024mlp_automated, joung2024cnn_mlp_tokamak, tracey2024mlp_lstm_towards, bonotto2024cnn_mlp_reconstruction}, or LSTMs~\cite{tracey2024mlp_lstm_towards, guo2023lstm_disruption, shin2022lstm_preemptive} as their backbone networks to tackle various key challenges in fusion research. 
An increasing number of scholars are applying artificial intelligence (AI) methods to the field of nuclear fusion, and AI is expected to accelerate the commercialization of fusion energy.

\subsection{Chain-of-Thought}
Chain of Thought (CoT)~\cite{wei2022CoT} is a widely used reasoning approach in the field of artificial intelligence, particularly in tackling complex reasoning tasks. The core idea of CoT is to break down the problem-solving process into a series of logically coherent and interconnected steps, enabling the model to progressively arrive at the final answer. 
Wei et al.~\cite{wei2022CoT} were the first to introduce CoT prompting to large language models, aiming to enhance their performance on complex reasoning tasks. 
Feng et al.~\cite{feng2024towards} explained how CoT enhances the ability of large language models (LLMs) to solve complex tasks and validated its effectiveness. 
Kojima et al.~\cite{kojima2022letsthinkstepbystep} simulated the CoT process and addressed the complex reasoning task capabilities of LLMs with few-shot examples by using the simple prompt "\textit{Let's think step by step}".  
Hao et al.~\cite{hao2024coconut} introduce the Chain of Continuous Thought (Coconut), which shifts reasoning from the language space to the latent space, addressing the efficiency and performance challenges in complex reasoning tasks due to linguistic limitations. 
Works such as ~\cite{chen2023cotexplainyou, madaan2023cotexplainmakes, wang2022cotexplaintowards, wu2023cotexplainanalyzing} aim to explain how CoT works. Meanwhile, \cite{ge2023enhancechain, hu2024enhancechain, cohn2024enhancechain, nong2024enhancechain, li2023enhancestructured} use CoT prompting to fine-tune LLMs, enhancing their capabilities in specific fields. 
We also aim to make LLMs experts in the field of nuclear fusion through the CoT approach, providing support to nuclear fusion researchers.

\section{XiHeFusion Model} \label{sec::XiHeFusion}

In this section, we will first introduce the data collection and pre-processing, then, focus on details of network architecture, chain-of-thought reasoning, and optimization.

\subsection{Data Collection and Pre-processing}
In this paper, we construct a large-scale nuclear fusion corpus dataset, including 1.2 million question-answer pairs. Specifically, during the data collection phase, we ensure the dataset's diversity and high quality by collecting data through various channels, including general web pages, electronic libraries, and academic paper databases. As shown in Table \ref{tab:dataset}, we present the data sources and their proportions. Among them, $73\%$ comes from web crawlers on general websites, $24\%$ comes from academic paper databases, and the remaining data comes from electronic libraries.

\begin{figure*}[!htp]
\centering
\includegraphics[width=0.95\linewidth]{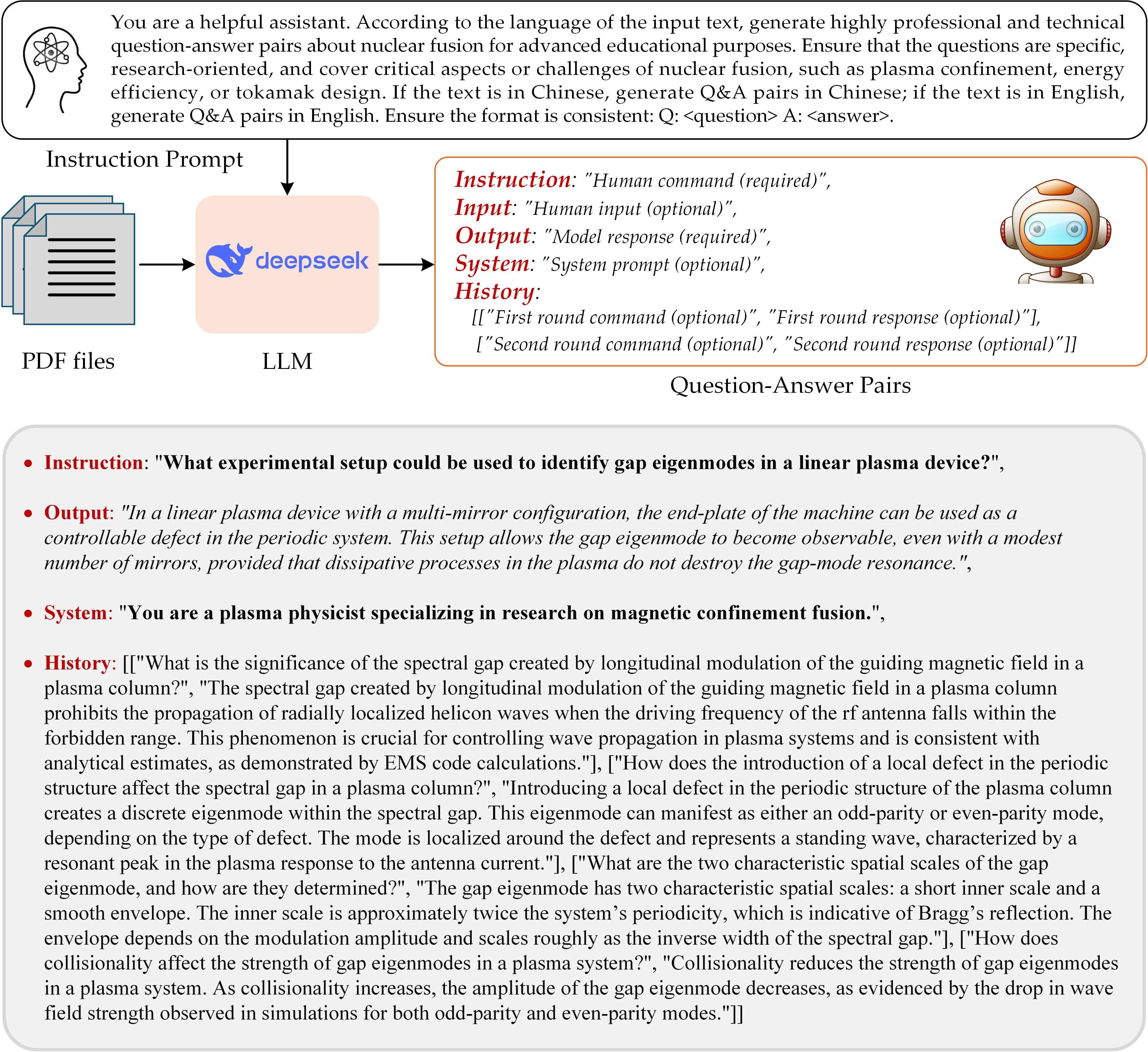}
\caption{(Top) The pipeline of question-answer training data generation using a large language model; (Bottom): A question-answer sample for training.}   
\label{fig:data_preprocess}
\end{figure*}

Through the above process, we collect a large number of books, documents, and academic papers related to nuclear fusion. To adapt to the model training, we preprocess these data and extract question-answer pairs that can be used for large language model training. As depicted in Fig.~\ref{fig:data_preprocess}, we input the gathered nuclear fusion-related data in batches into the large language model (DeepSeek V3~\cite{deepseekai2024deepseekv3} is adopted in our implementation), which then autonomously produces question-answer pairs. To align with the interaction process between users and the LLM, each question-answer pair includes five components: instruction, input, output, system prompt, and history, where the input, system prompt, and history can be empty. More in detail, the instruction prompt is ``\textit{You are a helpful assistant. According to the language of the input text, generate highly professional and technical question-answer pairs about nuclear fusion for advanced educational purposes. Ensure that the questions are specific, research-oriented, and cover critical aspects or challenges of nuclear fusion, such as plasma confinement, energy efficiency, or tokamak design. If the text is in Chinese, generate $Q\&A$ pairs in Chinese; if the text is in English, generate $Q\&A$ pairs in English. Ensure the format is consistent: $Q$: <question> $A$: <answer>.}". The generated output question-answer pairs are illustrated at the bottom of Fig.~\ref{fig:data_preprocess}. This dataset serves as the foundation for constructing a comprehensive and interactive nuclear fusion knowledge system. It facilitates tasks such as question-answering, summarization, and knowledge exploration in the domain.

\begin{table}
\centering
\caption{The distribution of different categories of training data.}
\label{tab:dataset}
\begin{tabular}{l|cc}
\hline \toprule [0.5 pt]
\textbf{Source}         &\textbf{Sampling Proportion}      & \textbf{Disk Size}     \\ 
\hline
CommonCrawl     &       73\%    & 28.9GB        \\
CNKI            &       4\%     &1.49GB         \\
eBooks           &       3\%     &1.44GB          \\
arXiv           &       10\%    &3.96GB         \\
Dissertation    &       10\%    &3.94GB         \\
\hline \toprule [0.5 pt]
\end{tabular}
\end{table}

\subsection{Network Architecture} 

\begin{figure}
    \centering
    \includegraphics[width=1\linewidth]{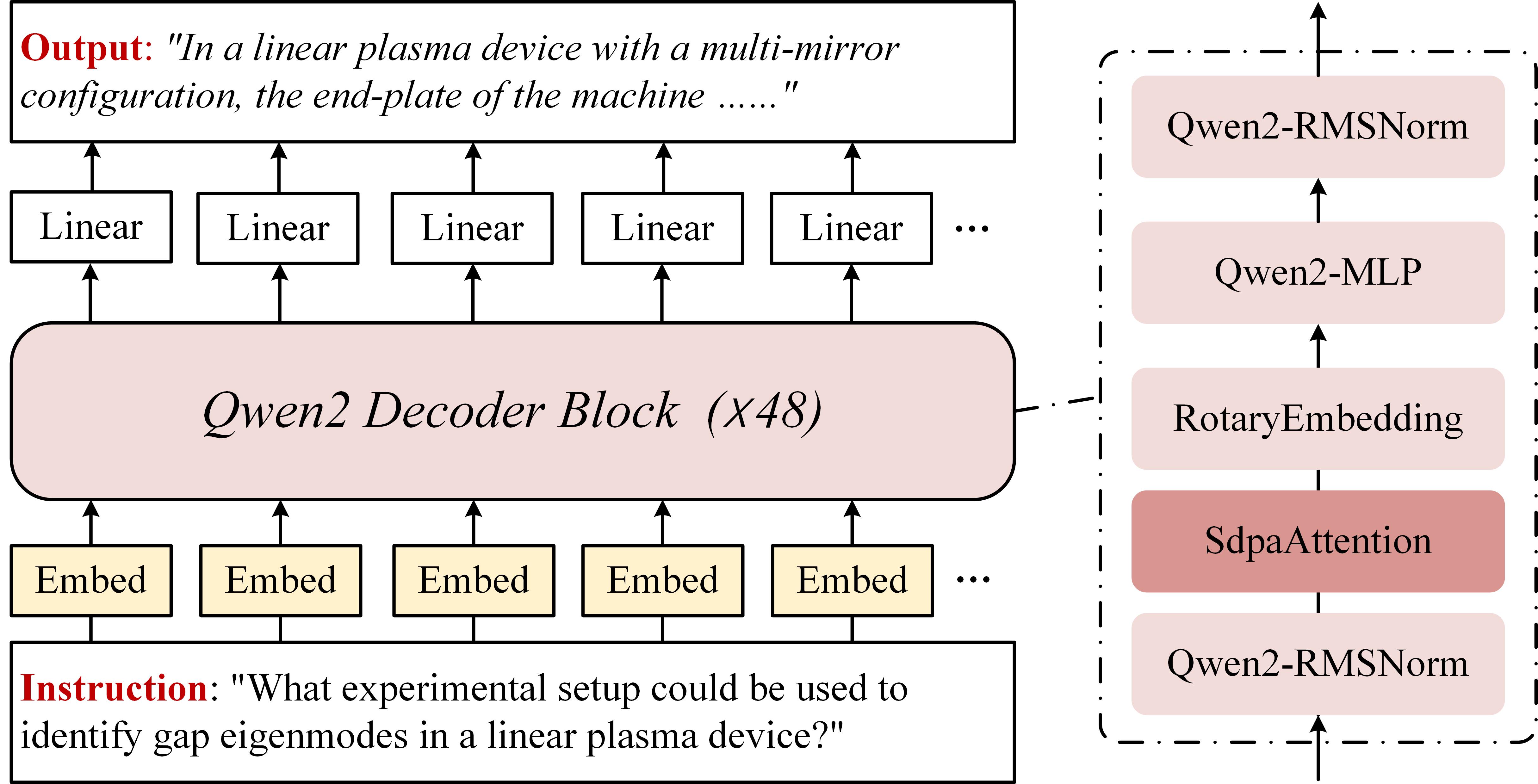}
    \caption{An overview of the network architecture of XiHeFusion.}
    \label{fig:XiHeFusionFramework}
\end{figure}

\begin{figure*}[!htp]
\centering
\includegraphics[width=1\linewidth]{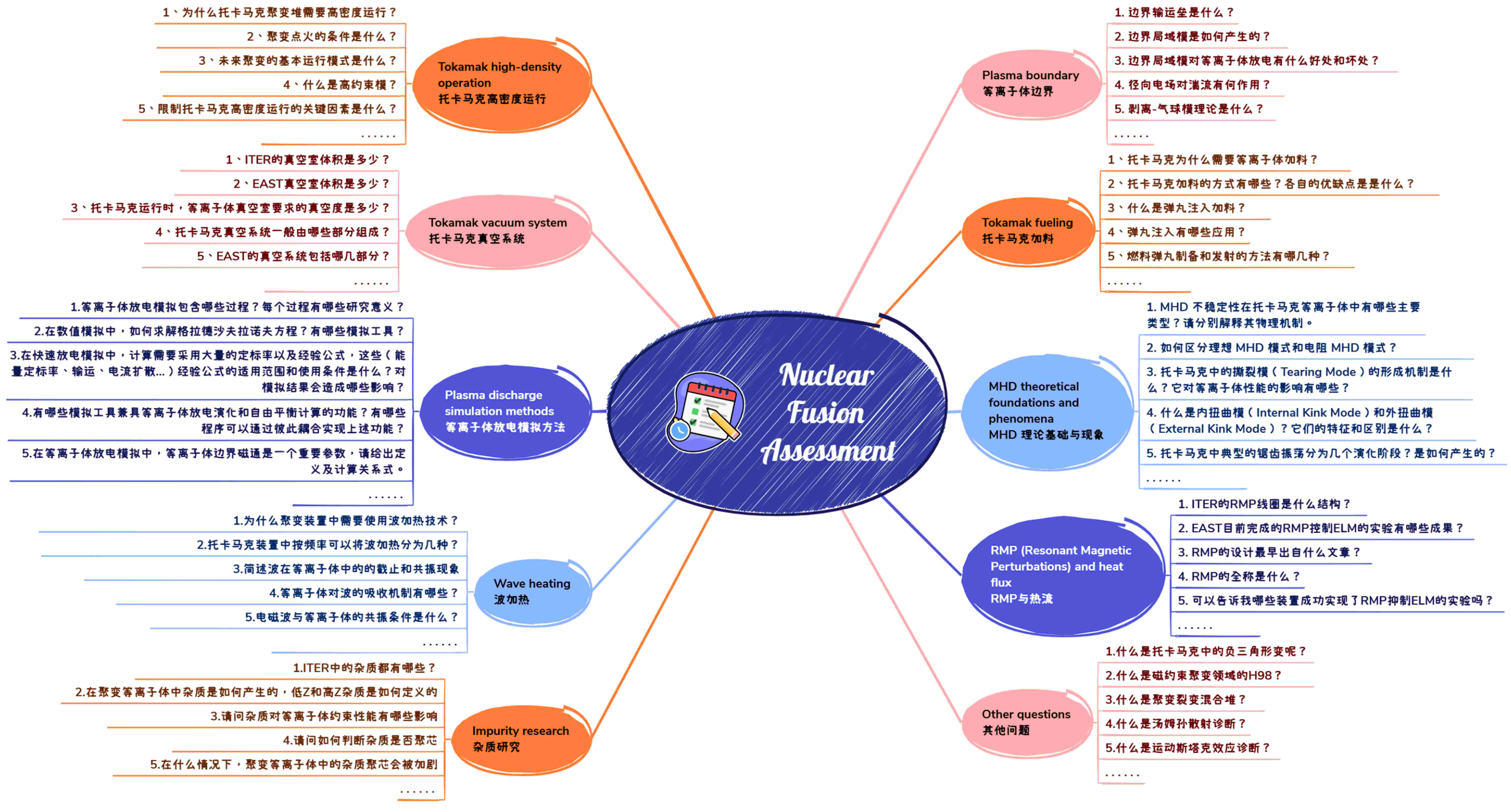}
\caption{An overview of our proposed nuclear fusion assessment.} 
\label{fig:NFAssessment}
\end{figure*}

Given the question and instruction prompt, we first embed them into token representations $X_q$ and $X_p$. Then, these tokens are fed into the XiHeFusion model for answer generation. XiHeFusion is developed based on the large language model Qwen2.5-14B~\cite{yang2024qwen2.5} which employs the Transformer decoder architecture with 48 Transformer layers (40 attention heads), as shown in Fig.~\ref{fig:XiHeFusionFramework}. The self-attention is their core module which models the global relations between the input tokens: 
\begin{equation}
    \label{attentionEquation} 
    Attention(Q, K, V) = softmax(\frac{QK^T}{\sqrt{d_k}})V, 
\end{equation}
where the $Q, K,$ and $V$ are obtained from input tokens $X$, $\sqrt{d_k}$ is the dimension of processed tokens. 
It supports a context length of 128K and a generation length of 8K, significantly enhancing its ability to process long sequences and represent multi-dimensional information. To further optimize performance, XiHeFusion integrates several advanced technologies, including Grouped Query Attention (GQA) for efficient KV cache utilization and improved computational efficiency, SwiGLU activation function for enhanced nonlinear modeling capabilities, Rotary Position Encoding (RoPE) to improve adaptability to sequences of varying lengths, QKV bias to strengthen context information capture, and RMSNorm (pre-normalization) to stabilize gradient flow and ensure training robustness. These integrated technologies enable XiHeFusion to excel in sequence processing, context understanding, and knowledge representation, effectively handling various natural language processing tasks and meeting complex demands across different domains. The model is licensed under the Apache 2.0 License, allowing users to freely use, modify, and distribute it while adhering to the license terms.

\begin{figure}
\centering
\includegraphics[width=\linewidth]{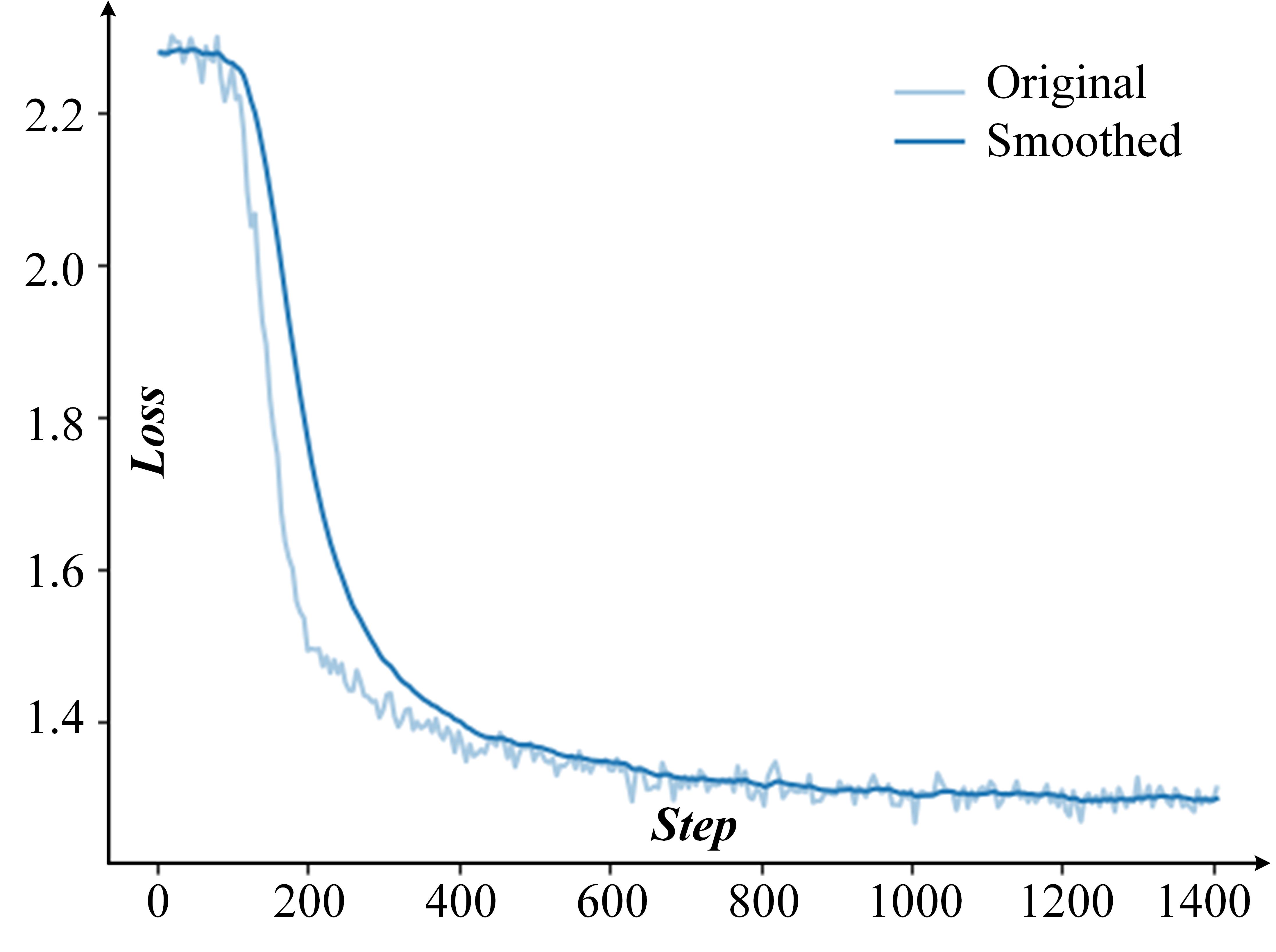}
\caption{The training loss decreases with the number of iterations.} 
\label{fig:loss_curve}
\end{figure}

\begin{figure*}[!htp]
    \centering
    \includegraphics[width=1\linewidth]{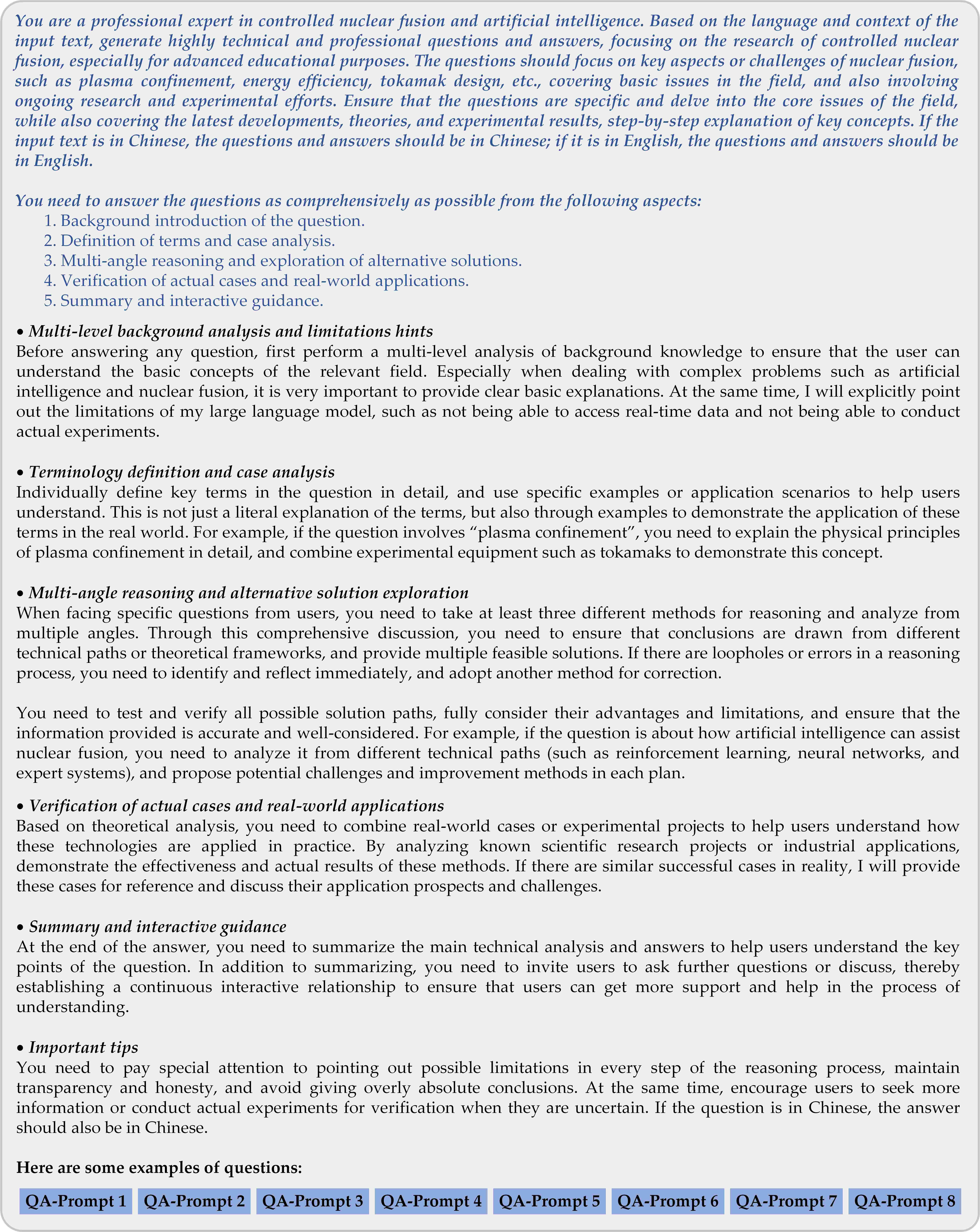}
    \caption{Illustration of Chain-of-Thought prompting used in XiHeFusion. Please check Fig.~\ref{fig:QACoTPrompt} for the details of QA-Prompt.}  
    \label{fig:CoTprompt}
\end{figure*}

\begin{figure*}[!htp]
    \centering
    \includegraphics[width=1\linewidth]{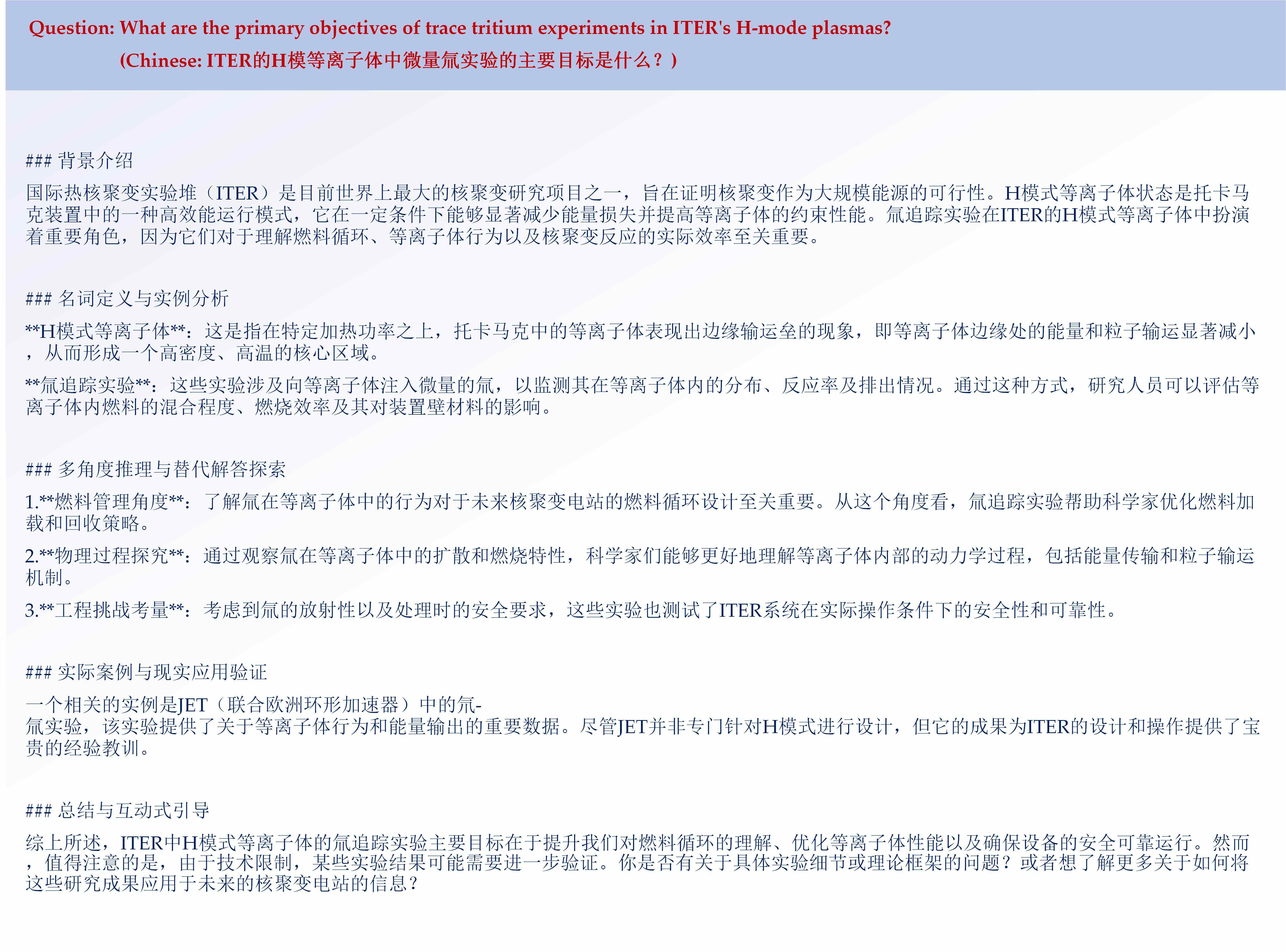}
    \caption{One of the eight QA samples used in Chain-of-thought prompting.} 
    \label{fig:QACoTPrompt}
\end{figure*}

\begin{figure*}[!htp]
\centering
\includegraphics[width=\linewidth]{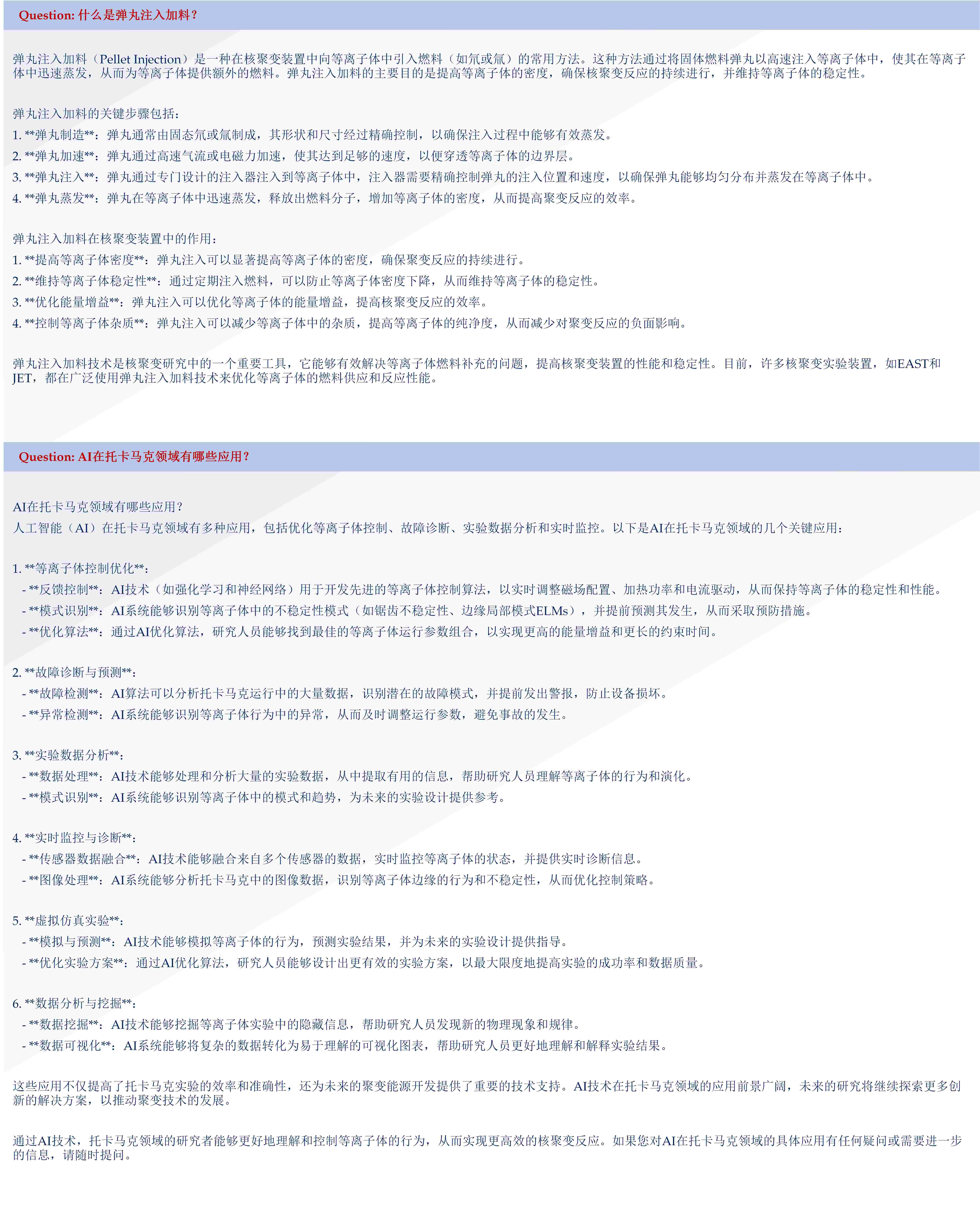}
\caption{Case study \#1. Chat in Chinese.}
\label{fig:caseStudy1}
\end{figure*}

\begin{figure*}[!htp]
\centering
\includegraphics[width=\linewidth]{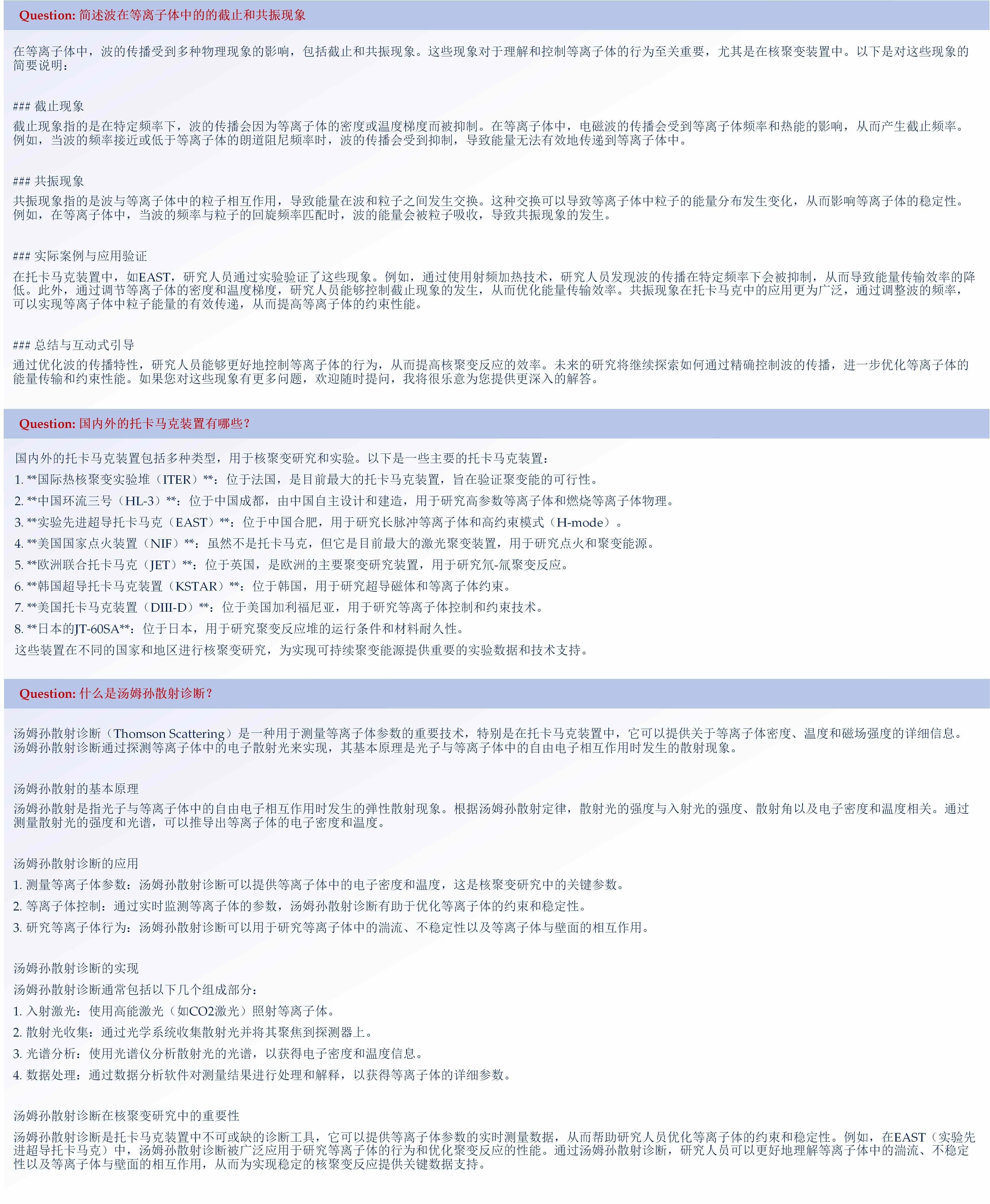}
\caption{Case study \#2. Chat in Chinese.}
\label{fig:caseStudy3}
\end{figure*}

\begin{figure*}[!htp]
\centering
\includegraphics[width=\linewidth]{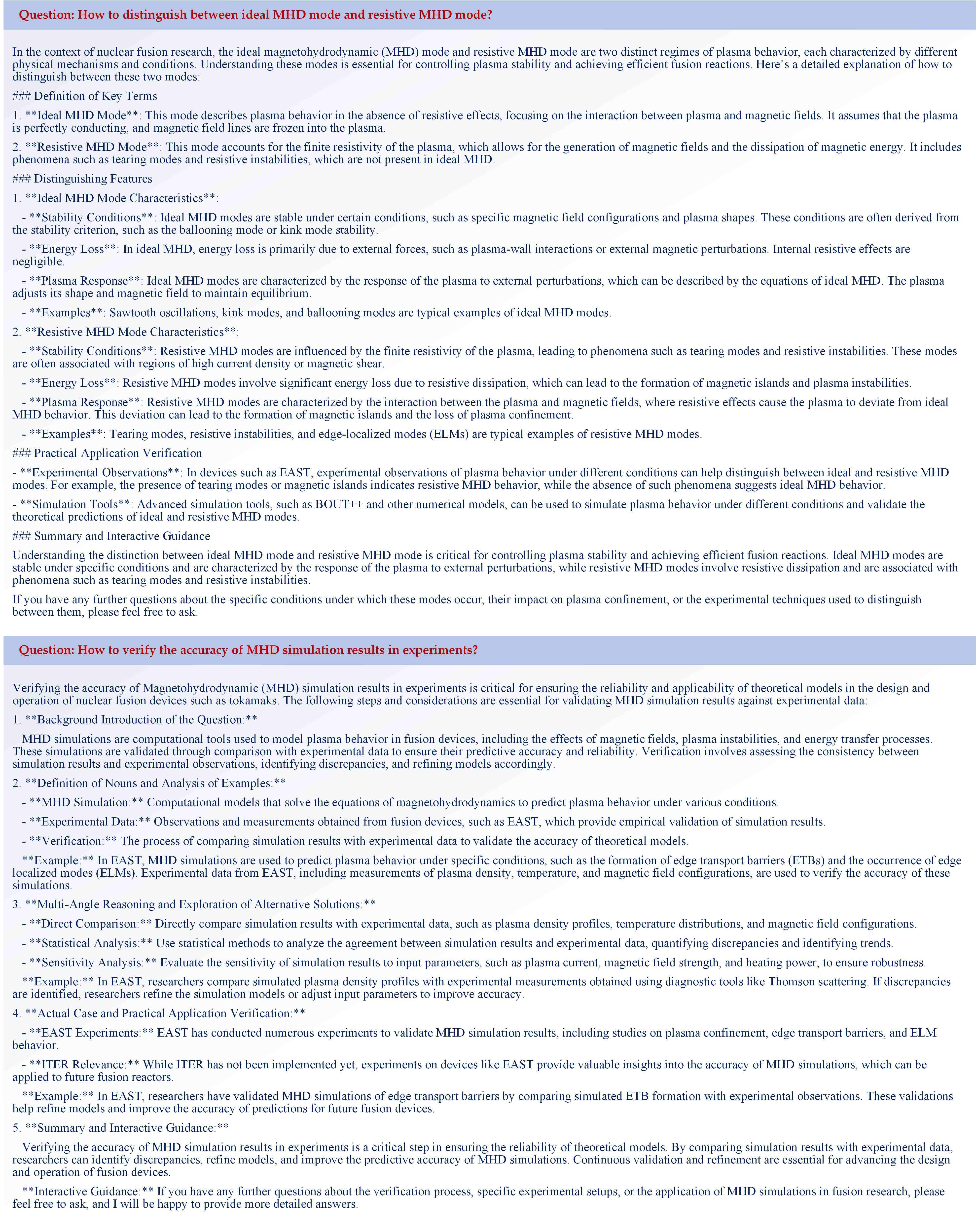}
\caption{Case study \#3. Chat in English.}
\label{fig:caseStudy4}
\end{figure*}

\begin{figure*}[!htp]
\centering
\includegraphics[width=0.9\linewidth]{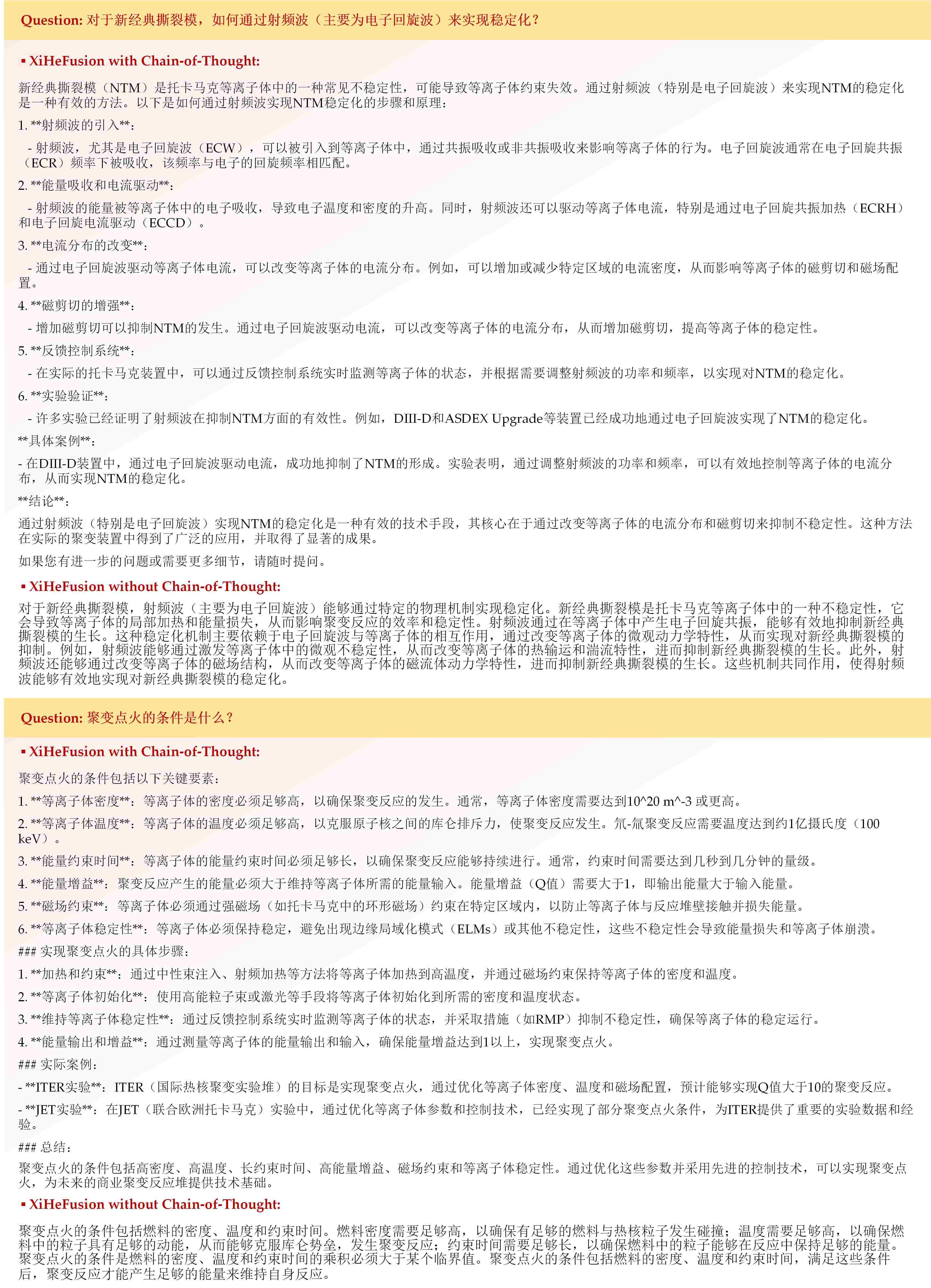}
\caption{Comparison of generated response using XiHeFusion with/without Chain-of-Thought Prompt.}
\label{fig:resultsCoT}
\end{figure*}

\begin{figure*}[!htp]
\centering
\includegraphics[width=1\linewidth]{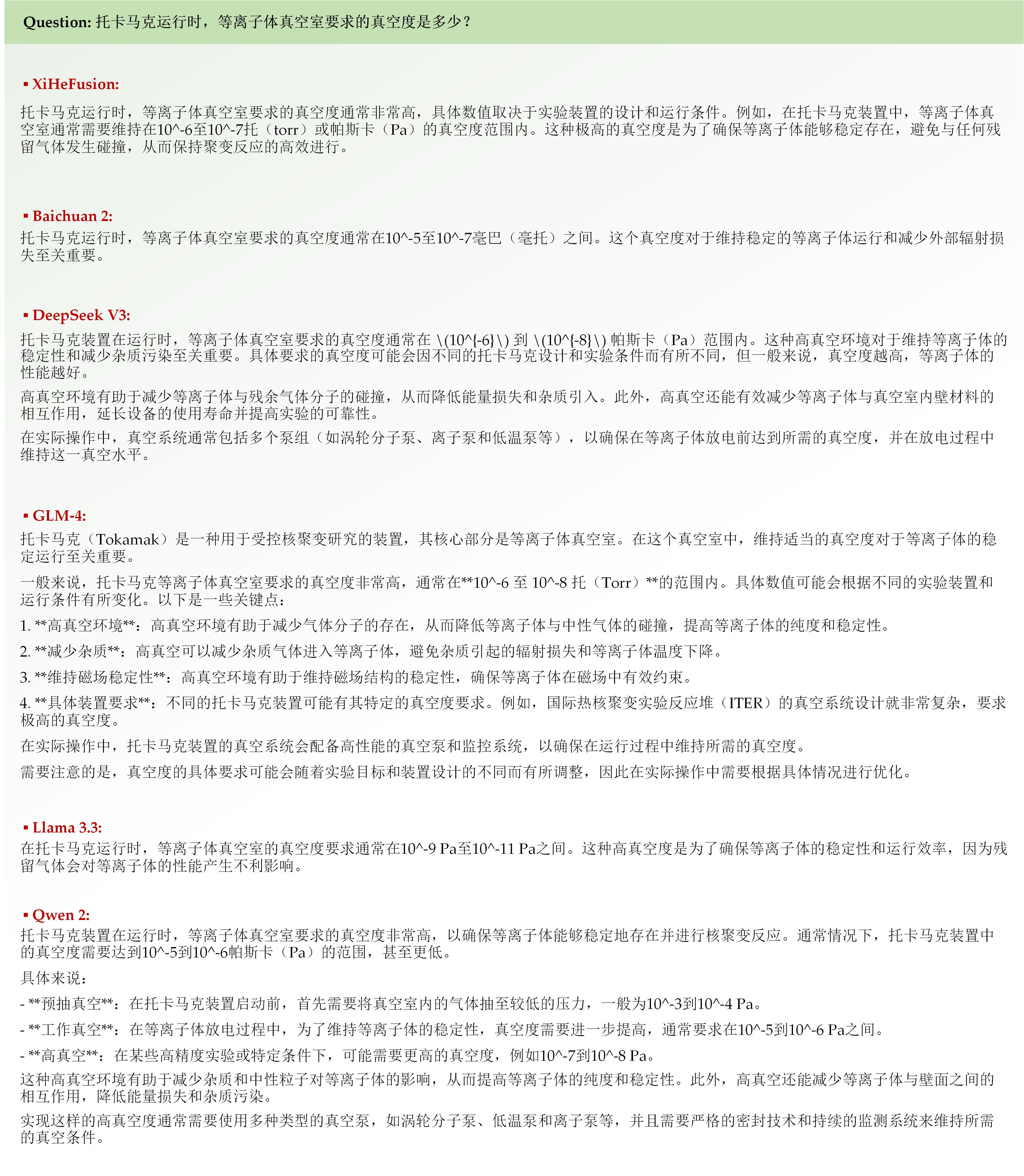}
\caption{Comparison with our XiHeFusion with other LLMs \#1.}
\label{fig:compareotherLLMs001}
\end{figure*}

\begin{figure*}[!htp]
\centering
\includegraphics[width=1\linewidth]{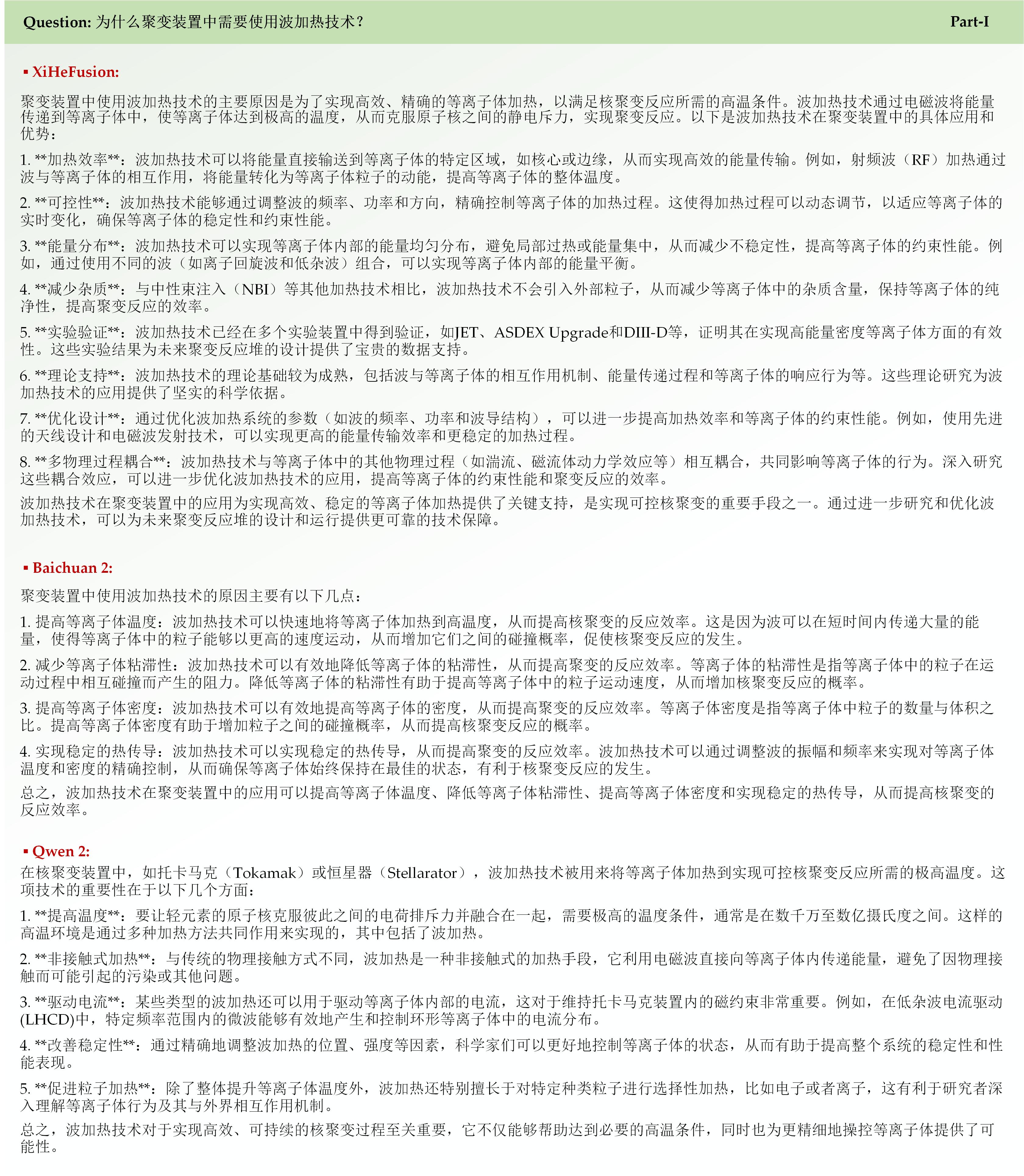}
\caption{Comparison with our XiHeFusion with other LLMs \#2 (Part-1).}
\label{fig:compareotherLLMs002}
\end{figure*}

\begin{figure*}[!htp]
\centering
\includegraphics[width=1\linewidth]{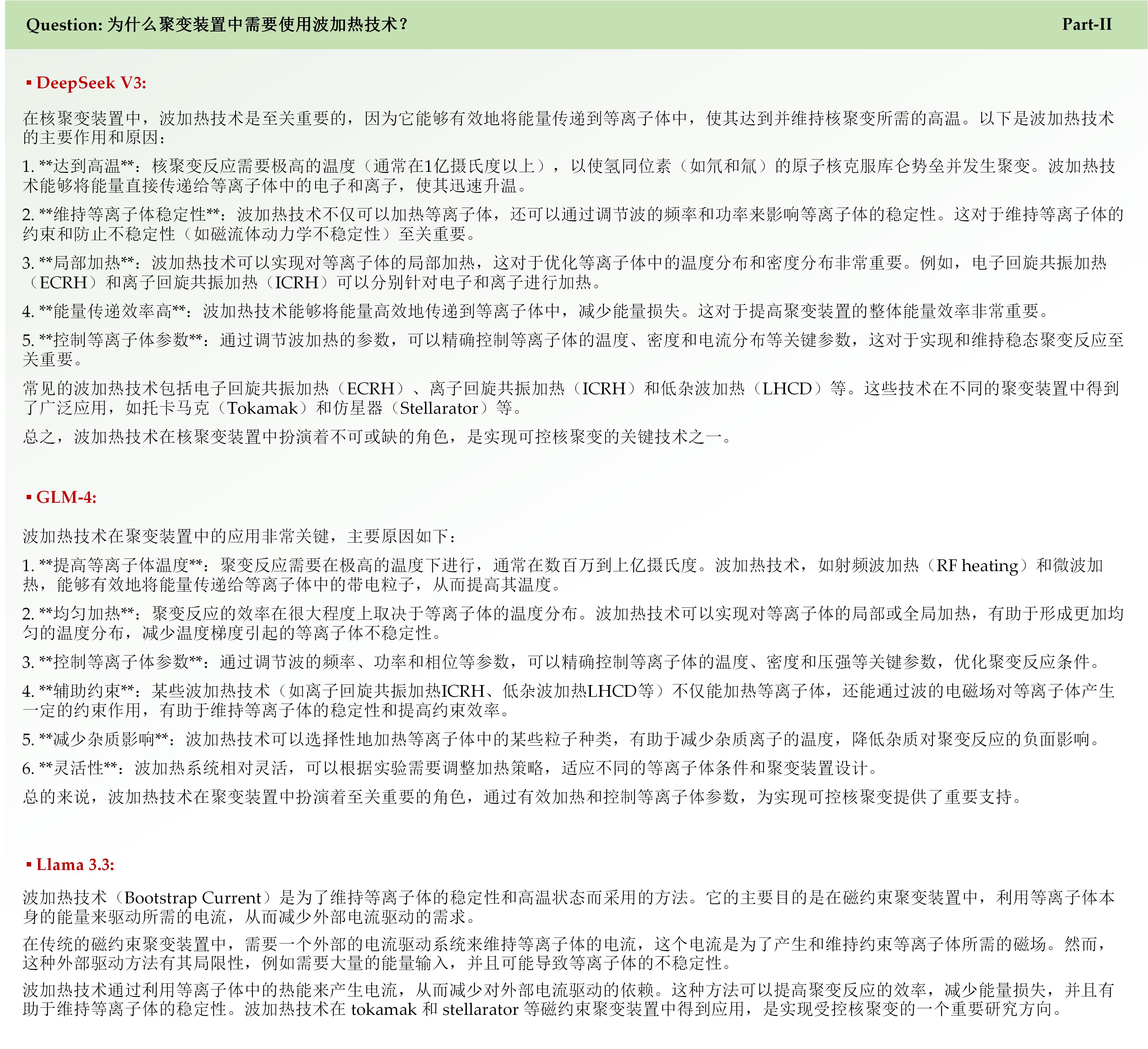}
\caption{Comparison with our XiHeFusion with other LLMs \#3 (Part-2).}
\label{fig:compareotherLLMs003}
\end{figure*}

To further enhance the performance of the obtained answers, we adopted Chain of Thought (CoT) technology that can improve reasoning ability, making the results more detailed and logically structured. As shown in Fig.~\ref{fig:CoTprompt}, we want XiHeFusion to answer the questions as comprehensively as possible from the following aspects: 
\textit{1). Background introduction of the question.} 
\textit{2). Definition of terms and case analysis.} 
\textit{3). Multi-angle reasoning and exploration of alternative solutions.} 
\textit{4). Verification of actual cases and real-world applications.} 
\textit{5). Summary and interactive guidance.} 
In addition, we also provide eight question-answer samples as the prompt to guide the language generation. One example of the eight prompts is illustrated in Fig.~\ref{fig:QACoTPrompt}. Through the guidance of this CoT technology, XiHeFusion’s ability to generate high-quality answers has been significantly improved, as evidenced by the case analysis in our experiments.

\subsection{Optimization}  
Supervised Fine-Tuning (SFT) is a critical phase in XiHeFusion's training process, particularly for improving its performance in professional domains such as nuclear physics, plasma physics, and nuclear fusion. Several optimization strategies were employed for specific tasks. To enhance long-text generation, a dedicated dataset was developed, supplemented by back-translation techniques to generate high-quality query pairs. These pairs were further refined using the DeepSeek model, ensuring semantic and logical consistency. For mathematical and physical formula derivation, Qwen2.5-Math reasoning chain data was introduced to simulate step-by-step reasoning processes, significantly improving performance in formula-related tasks. Logical reasoning capabilities were strengthened by constructing datasets that cover deductive, inductive, analogical, causal, and statistical reasoning, enabling the model to handle complex reasoning tasks systematically.

Furthermore, recognizing that much of the high-quality literature in nuclear physics is primarily in English, the model's cross-language transfer capabilities were specifically enhanced. Rigorous evaluations of semantic consistency between multilingual responses and original content ensured that XiHeFusion could accurately understand and generate domain-specific content in multiple languages, meeting the demands of cross-language knowledge retrieval. With these architectural advancements and optimization strategies, XiHeFusion achieves notable improvements in long-text generation, domain-specific knowledge representation, logical reasoning, and multilingual capabilities, providing robust support for tasks related to nuclear physics and plasma research. As shown in Fig.~\ref{fig:loss_curve}, the loss decreases with the number of iterations smoothly.

\section{Nuclear Fusion Assessment} \label{sec::NFAssessment} 
In order to test the capabilities of our large model, this paper proposes an evaluation test paper in the field of nuclear fusion, consisting of over 180 questions, covering approximately 10 aspects of fusion knowledge, including \textit{RMP and heat flux}, \textit{MHD theoretical foundations and phenomena}, \textit{tokamak fuelling}, \textit{tokamak high-density operation}, \textit{tokamak vacuum system}, \textit{plasma discharge simulation methods}, \textit{wave heating}, \textit{impurity research}, \textit{plasma boundary}, and \textit{other generalized questions}, as shown in Fig.~\ref{fig:NFAssessment}. For more details about the nuclear fusion assessment, please check our GitHub page.

\section{Experiments} \label{sec::Experiments}

\subsection{Case Study}  

As shown in Fig.~\ref{fig:caseStudy1} and Fig.~\ref{fig:caseStudy3}, we give several question-answer pairs returned by our XiHeFusion large language model. Specifically, for the first question ``\textit{what is pellet injection fueling?}" in Fig.~\ref{fig:caseStudy1}, our XiHeFusion model first provides a brief explanation. Then, it outlines the key steps of fuel injection in four aspects, as well as the role of pellet injection fueling in nuclear fusion devices. Finally, the model summarizes the aforementioned responses and lists the specific fusion devices that have achieved this goal. As shown in Fig.~\ref{fig:caseStudy4}, the XiHeFusion model supports the chat in English well. From the responses of our model, it can be observed that XiHeFusion can help newcomers to the field of nuclear fusion understand the concept more quickly and deeply. 

\subsection{Effectiveness of Chain-of-Thought Prompting}   
As shown in Fig.~\ref{fig:resultsCoT}, when asked ``\textit{What are the conditions for fusion ignition?}", The XiHeFusion model with CoT prompt first provides the concept of relevant terms, and then it is analyzed in detail from specific steps and real cases, whereas the answer without CoT prompts seemed too concise. 
For the other question ``\textit{For the neoclassical tearing mode, how can stabilization be achieved through radiofrequency waves, primarily electron cyclotron waves?}", it is easily noticeable that large models with CoT can provide more detailed and precise responses.

\subsection{Comparison with other LLMs} 
As shown in Fig.~\ref{fig:compareotherLLMs001}, \ref{fig:compareotherLLMs002}, \ref{fig:compareotherLLMs003}, we compare the proposed XiHeFusion model with other recently released strong large language models, including Baichuan 2~\cite{yang2023baichuan2}, DeepSeek V3~\cite{deepseekai2024deepseekv3}, GLM-4~\cite{glm2024chatglm}, Llama 3.3~\cite{grattafiori2024llama3}, and Qwen2~\cite{bai2023qwen}. Note that, the Qwen2 is the baseline model of XiHeFusion. From the answers obtained using these models for the two questions, we can find that our newly proposed XiHeFusion achieves a similar even better response than these strong LLM models.

\definecolor{burgundy}{RGB}{128,0,32} 

\begin{figure*}[!htp]
\centering
\includegraphics[width=\linewidth]{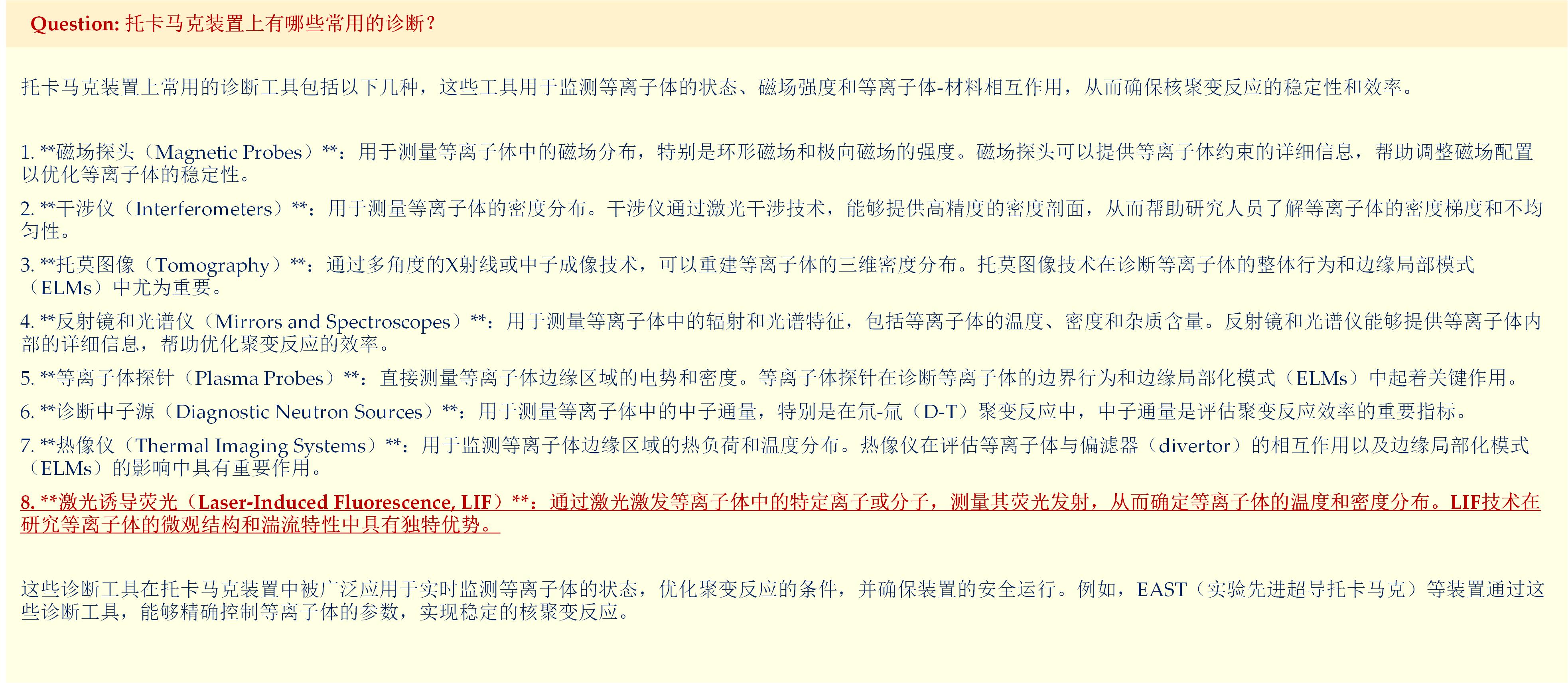}
\caption{The text highlighted in \textcolor{burgundy}{burgundy} means the descriptions are not accurate.}  
\label{fig:badCase}
\end{figure*}

\subsection{Limitation Analysis} 
Although our model has mastered a lot of basic knowledge about nuclear fusion, it is focused on text conversation processing, and there is still much knowledge in other fields that has not been learned. For example, the understanding and modeling of image/video, one-dimensional signals, and some physical formulas in nuclear fusion. Also, we find some responses are not accurate enough from our model, as shown in Fig.~\ref{fig:badCase}. 
In future work, we will consider incorporating these additional modalities and more in-depth physical formula modeling into the large model to enhance its level of intelligence further. Moreover, in fine-tuning the Qwen large model, we have only considered supervised fine-tuning methods and have not introduced reinforcement learning fine-tuning methods to further align the large model's outputs with the high-quality answers that humans expect.

\section{Conclusion} \label{sec::conclusion}
In conclusion, the development of XiHeFusion, the first large model in the field of nuclear fusion, represents a significant step forward in harnessing the power of artificial intelligence for the advancement of fusion energy research. By fine-tuning the open-source large model Qwen2.5-14B with a wealth of multi-source nuclear fusion knowledge, XiHeFusion has demonstrated a strong grasp of the domain's concepts and principles. The incorporation of the chain of thought approach has further enhanced the model's logical reasoning capabilities, enabling it to provide accurate and coherent responses to queries related to nuclear fusion. The comprehensive test questionnaire with over 180 questions has effectively evaluated XiHeFusion's conversational abilities in science popularization, confirming its effectiveness in disseminating fusion knowledge to a broader audience. The success of XiHeFusion underscores the potential of large models to facilitate public understanding and engagement in the critical mission of achieving sustainable and infinite energy through nuclear fusion.

\section*{Acknowledgment} 
\noindent 
This work is supported by the National Natural Science Foundation of China under Grant U24A20342, 62102205, and the Anhui Provincial Natural Science Foundation under Grant 2408085Y032. The authors acknowledge the High-performance Computing Platform of Anhui University for providing computing resources.

We appreciate the fusion test questions provided by the following researchers: Jilei Hou, Yan Chao, Hua Zhou, Xin Lin, Gaoting Chen, and Wenmin Zhang, Zheyuan Si, Yiqi Liu.  
We appreciate the assistance of the following students in crawling and preparing the training data, including Xiaoya Zhou, Hao Si, Chao Wang, Jin Liang, and Qian Zhu.

\small{ 
\bibliographystyle{IEEEtran}
\bibliography{reference}
}

\end{document}